\documentclass{article}

\usepackage{iclr2027_conference,times}
\usepackage[T1]{fontenc}

\usepackage{algorithm}
\usepackage{algpseudocode}
\usepackage{graphicx}
\graphicspath{{figures/}{./}}
\usepackage{amsmath}
\usepackage{amssymb}
\usepackage{amsfonts}
\usepackage{bm}
\usepackage{booktabs}
\usepackage{nicefrac}
\usepackage{microtype}
\usepackage{xcolor}
\usepackage{enumitem}
\usepackage{multicol}
\usepackage{multirow}
\usepackage{makecell}
\usepackage{colortbl}
\usepackage{threeparttable}
\usepackage{url}

\definecolor{iclrblue}{rgb}{0.21,0.49,0.74}
\definecolor{oursbg}{RGB}{240,240,240}
\usepackage[
    pagebackref,
    breaklinks,
    colorlinks,
    allcolors=iclrblue
]{hyperref}
\usepackage[capitalize,noabbrev]{cleveref}

\crefname{section}{Sec.}{Secs.}
\Crefname{section}{Section}{Sections}
\crefname{table}{Tab.}{Tabs.}
\Crefname{table}{Table}{Tables}
\crefname{figure}{Fig.}{Figs.}
\Crefname{figure}{Figure}{Figures}

\newcommand{\best}[1]{\textcolor{red}{#1}}
\newcommand{\second}[1]{\textcolor{blue}{#1}}
\newcommand{\third}[1]{\textcolor{green!60!black}{#1}}

\newcommand{\safeincludegraphics}[2][]{%
  \IfFileExists{figures/#2}{%
    \includegraphics[#1]{figures/#2}%
  }{%
    \IfFileExists{#2}{%
      \includegraphics[#1]{#2}%
    }{%
      \fbox{\parbox[c][0.20\textheight][c]{0.92\linewidth}{\centering
        Missing figure file\\[0.5em]
        \texttt{\detokenize{#2}}\\[0.5em]
        Upload this file to the \texttt{figures/} folder.}}%
    }%
  }%
}

\newcommand{\ablgain}[1]{\textcolor{teal}{#1}}
\newcommand{\ablloss}[1]{\textcolor{red!70!black}{#1}}

\title{Recurrent Second-Order Learning for Efficient Video Super-Resolution}
\iclrfinalcopy

\author{%
  \parbox[t]{\dimexpr\textwidth-2\tabcolsep\relax}{%
    \centering\normalfont\small
    Jintong Hu$^{1*}$, Bin Chen$^{2*}$, Zhenyu Hu$^{1}$,
    Jiayue Liu$^{2}$, Guo Wang$^{1}$, Lu Qi$^{13\dagger}$\\[2pt]
    $^{1}$Insta360 Research \quad
    $^{2}$Peking University \quad
    $^{3}$Wuhan University
  }%
}

\begin{document}

\maketitle
\fancyhead{}
\renewcommand{\headrulewidth}{0pt}

\begin{abstract}
Video super-resolution (VSR) seeks to reconstruct high-resolution frames from low-resolution inputs. 
While diffusion-based methods have substantially improved perceptual quality, extending them to video remains challenging for two reasons: strong generative priors can introduce temporal instability, and multi-frame diffusion pipelines are often too expensive for practical deployment.
To address both challenges simultaneously, we propose InstaVSR, a lightweight diffusion framework for efficient video super-resolution.
InstaVSR combines three ingredients: (1) a pruned one-step diffusion backbone that removes several costly components from conventional diffusion-based VSR pipelines, (2) recurrent training with flow-guided temporal regularization to improve frame-to-frame stability, and (3) dual-space adversarial learning in latent and pixel spaces to preserve perceptual quality after backbone simplification.
On an NVIDIA RTX 4090, InstaVSR processes a 30-frame video at 2K$\times$2K resolution in under one minute with only 7 GB of memory usage, substantially reducing the computational cost compared to existing diffusion-based methods while maintaining favorable perceptual quality with significantly smoother temporal transitions.
\end{abstract}

\section{Introduction}
\label{sec:intro}

Efficient video super-resolution (VSR) aims to reconstruct high-resolution (HR) video frames from low-resolution (LR) observations while maintaining practical computational cost. It supports a wide range of applications, including video streaming~\citep{liu2017robust,zhang2020improving,zhang2021video}, surveillance enhancement~\citep{farooq2021human,liu2024advancing,ding2020deep}, film restoration~\citep{berlincioni2025high,peng2025mitigating}, and medical imaging and related reconstruction tasks~\citep{wang2022application,hu2024residual,cammarasana2023super}.

Despite the great success achieved by existing VSR methods through lightweight architectures~\citep{fuoli2019efficient,isobe2020video}, recurrent propagation~\citep{sajjadi2018frame,yan2019frame,chan2021basicvsr,liang2022recurrent}, and multi-frame alignment~\citep{wang2019edvr,chan2022basicvsr++}, these techniques are typically studied in isolation.
Moreover, when diffusion models are adopted for efficient one-step VSR, their strong generative priors may introduce temporal flickering by synthesizing inconsistent details across frames.
Thus, one question is raised: \emph{how can we systematically unify these designs within a diffusion framework to achieve a better efficiency--quality trade-off while balancing spatial synthesis and temporal consistency?}

To address this issue, we systematically revisit the design space of efficient video super-resolution.
By an extensive analysis of over 10 core design components, we propose InstaVSR following two key principles including temporal consistency and spatial efficiency.
Specifically, InstaVSR is a simple multi-input single-output (MISO) framework with lightweight flow-based alignment among frames, avoiding costly spatiotemporal 3D modeling.
Then, we streamline the diffusion backbone by removing redundant VAE and text-conditioning components and adopting one-step generation, while dual-space adversarial supervision preserves its generative capability.

Even with an optimized module combination identified through extensive ablations, we observe that InstaVSR still exhibits unstable synthesis across frames under varying noise levels.
We attribute this limitation to first-order learning, which directly learns the mapping from LR observations to HR targets, whereas stable reconstruction may require iterative refinement over multiple rounds to progressively converge.
Motivated by this observation, we introduce recurrent second-order learning, which feeds previous SR predictions back as new observations for subsequent reconstruction.
Given a sliding window of aligned LR frames, each predicted SR frame is written back to the temporal buffer to replace its LR counterpart and participate in subsequent predictions, allowing previously super-resolved frames to be revisited within an SR-enhanced context to yield second-order SR predictions.
Furthermore, we couple this recurrent process with flow-guided temporal regularization to explicitly enforce the consistency of motion-aligned predictions across neighboring frames.

At last, extensive experiments on multiple VSR benchmarks demonstrate that InstaVSR achieves a favorable balance among efficiency, perceptual quality, and temporal consistency through systematic module integration and recurrent second-order learning.
In particular, the current implementation contains only 0.45B parameters and processes a 30-frame sequence at 2K×2K resolution in approximately 0.77 minutes on an RTX 4090, with only 7 GB of GPU memory, demonstrating the practical efficiency of the proposed design.

\begin{figure}[t]
    \centering
    \includegraphics[width=\linewidth]{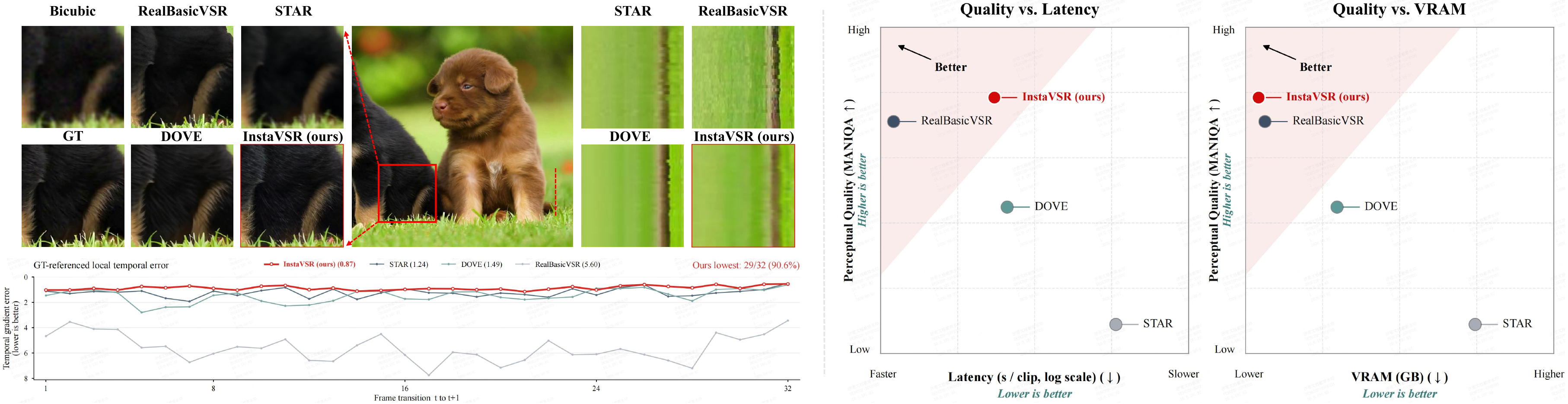}
    \caption{
        \textbf{InstaVSR delivers detailed and temporally consistent video super-resolution at low computational cost.} On the illustrated sequence, InstaVSR reconstructs fine spatial details and achieves the lowest GT-referenced local temporal gradient error on 90.6$\%$ of frame transitions. It also offers higher perceptual quality with lower inference latency and memory consumption.
    }
    \label{fig:teaser}
\end{figure}

Our contributions can be summarized as follows:
\begin{itemize}[leftmargin=*]
\item We investigate and optimize key components for efficient VSR, including a lightweight one-step backbone, aligned multi-frame observations, simplified conditioning, and dual-space adversarial learning, and integrate them into a unified framework InstaVSR.
\item Beyond conventional LR-to-SR supervision, we introduce recurrent second-order learning that progressively feeds first-order SR predictions back as observations for second-order reconstruction. This recurrent refinement exposes prediction errors and promotes temporally stable details without introducing heavy temporal architectures.
\item Extensive experiments on standard VSR benchmarks demonstrate that InstaVSR achieves competitive perceptual quality and temporal consistency with substantially lower memory consumption and inference latency than existing diffusion-based VSR methods.
\end{itemize}

\section{Related Work}
\label{sec:related_work}

\begin{figure}[t]
  \centering
  \safeincludegraphics[width=1\linewidth]{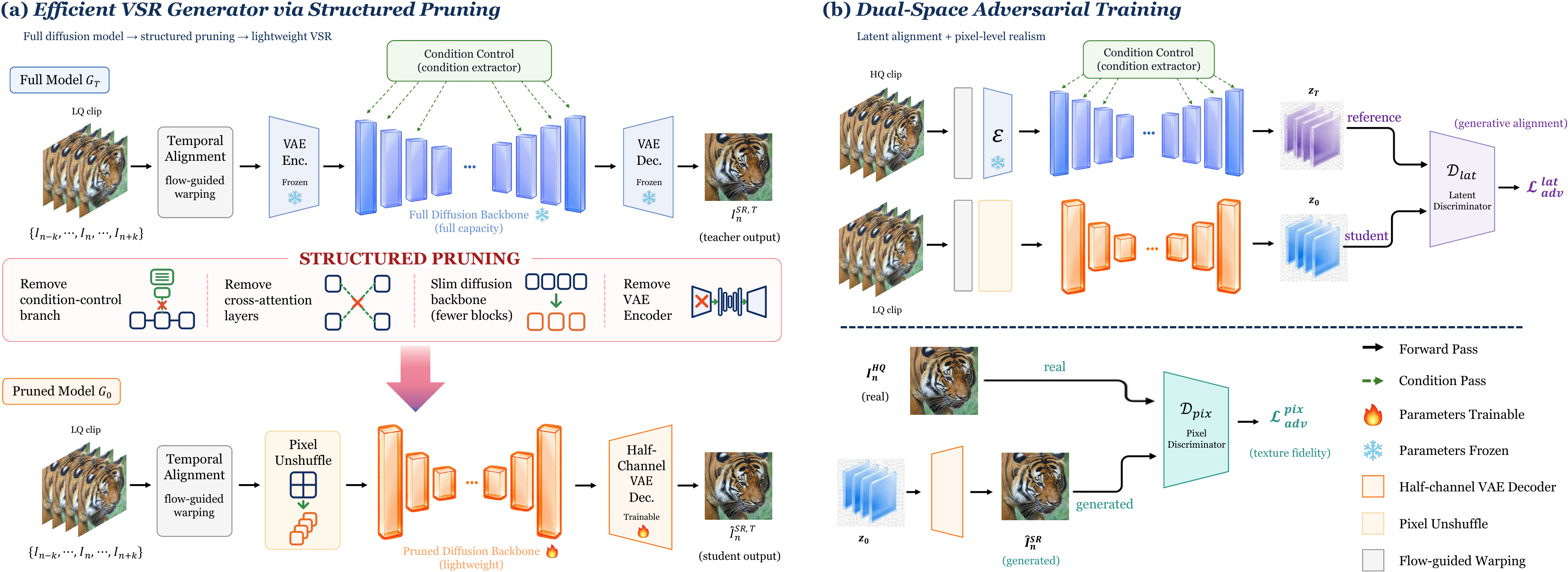}
   \caption{\textbf{Overview of Our Pruned Video Super-Resolution Framework and Dual-Space Adversarial Training Strategy.} (a) We prune the base model by removing the conditional control module for efficiency. A multi-frame input, single-frame output architecture enhances temporal stability, eliminating the need for the 3D VAE structure. (b) Our dual-space adversarial training uses latent and pixel discriminators. The latent discriminator aligns the pruned model's generative capacity with the full model, while the pixel discriminator enhances texture fidelity and prevents over-generation of fine details in reconstructed frames.}
   \label{fig:2}
\end{figure}

Efficient VSR must balance two objectives: exploiting
neighboring frames without temporal flickering or recurrent error
accumulation, and keeping spatial reconstruction tractable at high
resolution. Accordingly, we review prior work on temporal consistency
and spatial efficiency.

\subsection{Temporal Consistency in Video Super-Resolution}

Temporal consistency is commonly achieved through frame alignment,
recurrent propagation, or explicit temporal modeling.
Early frame-recurrent methods reuse reconstructed frames or features for
temporal coherence and efficiency.
FRVSR~\citep{sajjadi2018frame} propagates the previously estimated HR
frame to reconstruct the next frame, while FFCVSR~\citep{yan2019frame}
combines local multi-frame information with previously reconstructed
frames and features.
RSDN~\citep{isobe2020video} decomposes recurrent representations into
structure and detail components for robustness to appearance changes and
accumulated errors.
EDVR~\citep{wang2019edvr} instead uses deformable alignment and
temporal-spatial attention for neighboring-frame aggregation.
BasicVSR~\citep{chan2021basicvsr} propagates features bidirectionally,
while BasicVSR++~\citep{chan2022basicvsr++} introduces second-order grid
propagation and flow-guided deformable alignment.
RVRT~\citep{liang2022recurrent} combines recurrent propagation with local
attention for long-sequence restoration.
Despite exploiting temporal redundancy, these methods may accumulate
prediction and alignment errors under complex motion and diverse real-world
degradation patterns.

Generative VSR adds a challenge because independently synthesized
high-frequency details may vary between adjacent frames.
Frame-wise methods such as StableSR~\citep{wang2024exploiting} and
DiffBIR~\citep{lin2024diffbir} provide strong perceptual priors but lack
explicit temporal correspondence.
Upscale-A-Video~\citep{zhou2024upscale} incorporates temporal layers and
flow-guided latent propagation, whereas MGLD-VSR~\citep{yang2024motion}
uses motion-guided latent diffusion.
These methods improve temporal coherence, but temporal layers, latent
propagation, and iterative diffusion sampling raise the computational cost of inference.

Our recurrent second-order learning differs from conventional feature
propagation.
Rather than propagating only hidden features, we write reconstructed SR
frames back to the observation buffer for reuse.
A target frame is first reconstructed using the current buffer,
then revisited after additional neighboring SR predictions
have been written back.
Here, ``second-order'' denotes reconstruction order, not the second-order
feature propagation of BasicVSR++~\citep{chan2022basicvsr++}.
Flow-guided regularization further constrains motion-aligned outputs
without heavy 3D temporal modules.

\subsection{Spatial Efficiency in Generative Super-Resolution}

Spatial efficiency is critical because VSR computation and memory grow
rapidly with frame resolution.
Conventional methods reduce these costs through compact operators or
efficient temporal propagation.
RLSP~\citep{fuoli2019efficient} propagates temporal information through
a compact recurrent latent state.
ECBSR~\citep{zhang2021edge} uses structural re-parameterization, while
ShuffleMixer~\citep{sun2022shufflemixer} uses lightweight channel-mixing
operations.
SwinIR~\citep{liang2021swinir} introduces window-based attention for
image restoration, and VRT~\citep{liang2024vrt} extends transformer
modeling to video restoration.
Although more efficient than large generative models, their
regression-oriented objectives tend to favor distortion metrics and
offer limited ability to synthesize realistic high-frequency details in reconstructed video frames.

Diffusion models provide stronger generative priors but incur substantial
spatial overhead.
Latent diffusion reduces feature resolution through a
VAE~\citep{rombach2022high}, while StableSR~\citep{wang2024exploiting}
and DiffBIR~\citep{lin2024diffbir} reuse pretrained diffusion priors for
detail generation.
However, iterative denoising and auxiliary conditioning remain expensive
for individual video frames.
Video-native diffusion architectures also use 3D VAEs, temporal
attention, or spatiotemporal transformers to process clips, further
increasing high-resolution memory consumption
\citep{yang2024cogvideox,brooks2024video,zheng2024open,lin2024open}.

One-step generation offers a more efficient alternative.
AdcSR~\citep{chen2025adversarial} shows that an SD2.1-based image
super-resolution model produces perceptually realistic results in one
forward pass, while AdcVSR~\citep{chenimproved} explores efficient video
diffusion through teacher--student distillation.
Nevertheless, directly extending an image model to video does not resolve
the cost of multi-frame spatial processing or the temporal instability of
synthesized details.


\section{Methodology}
\label{sec:method}

\subsection{Problem Formulation and Overview}

Given an LR video
$V^l\in\mathbb{R}^{T\times H_l\times W_l\times 3}$, we aim to recover its
corresponding HR video
$V^h\in\mathbb{R}^{T\times H_h\times W_h\times 3}$, where $T$ is the number
of frames, $H_l$ and $W_l$ are the height and width of each LR frame, and
$H_h=rH_l$ and $W_h=rW_l$ for an upscaling factor $r$.

Our goal is to learn a VSR model $G_\theta$ that reconstructs
$\hat{V}^h$ from $V^l$. To achieve efficient and temporally consistent
reconstruction, InstaVSR combines a lightweight multi-frame one-step
diffusion backbone, recurrent second-order learning, and dual-space
adversarial supervision.

\subsection{Basic Framework Design}

As illustrated in Figure~\ref{fig:2}, InstaVSR adopts a multi-input
single-output framework. For each target LR frame $I_t^l$, we select
$2k+1$ neighboring frames and align them to the target coordinate system:
\begin{equation}
    \tilde{I}_{t+i}^l
    =
    \mathcal{W}
    \left(
        I_{t+i}^l,
        F_{t+i\rightarrow t}
    \right),
    \qquad i\in[-k,k],
    \label{eq:frame_alignment}
\end{equation}
where $F_{t+i\rightarrow t}$ is the optical flow from frame $t+i$ to frame
$t$, and $\mathcal{W}$ denotes backward warping. The aligned frames are
concatenated and fed into the generator:
\begin{equation}
    \hat{I}_t^h
    =
    G_\theta
    \left(
        \operatorname{Concat}
        [\tilde{I}_{t-k}^l,\ldots,I_t^l,\ldots,\tilde{I}_{t+k}^l]
    \right).
    \label{eq:miso_generator}
\end{equation}

To reduce computation, we employ a pruned one-step diffusion backbone. The
VAE encoder in the generator input path is replaced by Pixel Unshuffle, while
text--image cross-attention, timestep embeddings, and redundant bottleneck
blocks are removed. During training, reconstructed frames are progressively
written back into the temporal buffer and reused in subsequent predictions.
This recurrent process enables second-order learning, while flow-guided
temporal regularization and dual-space adversarial supervision improve
temporal consistency and perceptual quality.

\subsection{Recurrent Second-Order Learning}
\label{sec:recurrent_training}

Frame-wise reconstruction losses supervise spatial quality but do not
explicitly constrain consistency across frames. We introduce recurrent
second-order learning that feeds SR predictions back into the observation
buffer and reconstructs an earlier frame using the updated context.
The resulting second-order prediction is paired with a neighboring
first-order prediction for flow-guided temporal supervision across neighboring reconstructed frames.

\begin{figure}[t]
    \centering
    \includegraphics[width=0.55\linewidth]{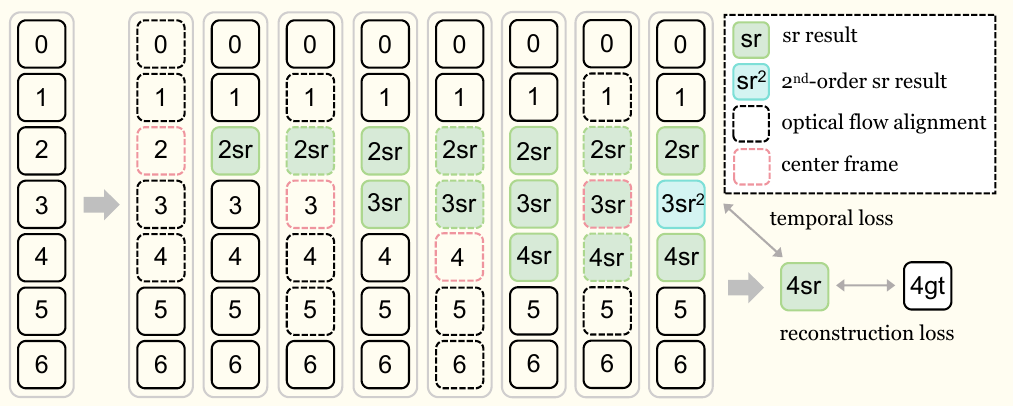}
    \caption{
    \textbf{Recurrent second-order training for VSR.}
    Frames 2, 3, and 4 are sequentially reconstructed from
    flow-aligned five-frame windows, with each SR prediction
    written back into the temporal buffer. Frame 3 is then
    reconstructed again using the updated context. Temporal
    loss compares its second-order result with the first-order
    result of frame 4 after flow alignment, while reconstruction
    loss supervises the latter against its ground truth.
    }
    \label{fig:3}
\end{figure}

\paragraph{Recurrent prediction.}
For the five-frame model, we sample seven consecutive LR frames
$\{I_{t-3}^l,\ldots,I_{t+3}^l\}$ and initialize a temporal buffer
with $B_j^{[0]}=I_j^l$. As illustrated in Figure~\ref{fig:3}, we
sequentially reconstruct frames $t-1$, $t$, and $t+1$ using their
aligned five-frame windows. Each SR prediction replaces its
corresponding observation before the next window is processed.
For $q=0,1,2$ and $j=t-1+q$, this process is written as
\begin{equation}
    \hat{I}_j^{h,(1)}
    =
    G_\theta\!\left(
        \mathcal{A}_j(B_{j-2:j+2}^{[q]})
    \right),
    \qquad
    B_j^{[q+1]}
    =
    \mathcal{R}\!\left(\hat{I}_j^{h,(1)}\right),
    \label{eq:recurrent_update}
\end{equation}
while all other buffer entries remain unchanged.
Here, $B^{[q]}$ denotes the buffer after $q$ updates,
$\mathcal{A}_j$ denotes flow-guided alignment, and
$\mathcal{R}$ converts an SR prediction to the representation
required by the input buffer. The superscript $(1)$ denotes
the first reconstruction of a frame, which may already use
previously reconstructed neighboring frames.

\paragraph{Second-order reconstruction.}
After generating $\hat{I}_{t+1}^{h,(1)}$, we revisit frame $t$
using the temporal buffer updated by preceding predictions:
\begin{equation}
    \hat{I}_t^{h,(2)}
    =
    G_\theta\!\left(
        \mathcal{A}_t(B_{t-2:t+2}^{[3]})
    \right).
    \label{eq:second_order}
\end{equation}
The updated window contains the first-order SR results of
frames $t-1$, $t$, and $t+1$. Reconstructing frame $t$ again
under this SR-enhanced context produces its second-order
prediction. In Figure~\ref{fig:3}, frames $t-1$, $t$, and $t+1$
correspond to frames 2, 3, and 4, respectively.
Here, ``second-order'' refers to reconstruction order, rather
than the second-order feature propagation used for recurrent feature aggregation in
BasicVSR++~\citep{chan2022basicvsr++}.

\paragraph{Reconstruction loss.}
We apply standard reconstruction supervision to the first-order
prediction of frame $t+1$ at the final sequential reconstruction step:
\begin{equation}
    \mathcal{L}_{\mathrm{rec}}
    =
    \ell_{\mathrm{rec}}\!\left(
        \hat{I}_{t+1}^{h,(1)}, I_{t+1}^h
    \right),
    \label{eq:reconstruction_loss}
\end{equation}
where $\ell_{\mathrm{rec}}$ denotes the reconstruction criterion
and $I_{t+1}^h$ is the corresponding HR target.
This is the supervision between the first-order SR result
of frame 4 and its ground truth in Figure~\ref{fig:3}.

\paragraph{Flow-guided temporal regularization.}
Temporal supervision compares the second-order prediction
$\hat{I}_t^{h,(2)}$ with the neighboring first-order prediction
$\hat{I}_{t+1}^{h,(1)}$. We estimate optical flow from the degraded
LR frames using a lightweight estimator distilled from
PWC-Net~\citep{sun2018pwc}, and align the SR predictions before
penalizing their appearance differences:
\begin{equation}
    \mathcal{L}_{\mathrm{temp}}
    =
    \frac{
        \sum_p \omega_{t\rightarrow t+1}(p)
        \rho\!\left(
            \hat{I}_{t+1}^{h,(1)}(p)
            -
            \mathcal{W}\!\left(
                \hat{I}_t^{h,(2)},
                F_{t\rightarrow t+1}
            \right)(p)
        \right)
    }{
        \sum_p \omega_{t\rightarrow t+1}(p)+\epsilon
    },
    \label{eq:temporal_loss}
\end{equation}
where $\mathcal{W}$ warps frame $t$ into the coordinate system
of frame $t+1$, and
$\rho(x)=\sqrt{x^2+\epsilon^2}$ is the Charbonnier penalty.
The confidence weight $\omega_{t\rightarrow t+1}$ incorporates
a visibility mask obtained through forward--backward flow
consistency to suppress unreliable correspondences.
This loss encourages agreement between neighboring predictions
generated at different reconstruction orders, complementing
the spatial supervision provided by $\mathcal{L}_{\mathrm{rec}}$.
Further implementation details are provided for the complete recurrent training procedure in
Appendix~\ref{app:recurrent_details}.

\begin{table}[t]
\centering
\caption{\textbf{Quantitative comparison between InstaVSR and other methods.}
The best, second-best, and third-best results of each dataset are marked in
\textcolor{red}{red}, \textcolor{blue}{blue}, and
\textcolor{green!60!black}{green}, respectively.}
\label{tab:main_results}
\renewcommand{\arraystretch}{1.12}
\setlength{\tabcolsep}{2.5pt}
\resizebox{\textwidth}{!}{
\begin{tabular}{c|l|ccc|ccc|cc}
\toprule
Dataset & Method
& LPIPS$\downarrow$
& DISTS$\downarrow$
& FID$\downarrow$
& NIQE$\downarrow$
& MANIQA$\uparrow$
& MUSIQ$\uparrow$
& $E_{\mathrm{warp}}\downarrow$
& DOVER$\uparrow$ \\
\midrule

~ & RealESRGAN \cite{wang2021real}
& 0.4339 & 0.2147 & 95.7204
& \third{3.472} & 0.6088 & 69.2294
& 3.1010 & \second{71.9639} \\

~ & RealViformer \cite{zhang2024realviformer}
& \second{0.2144} & 0.1500 & \second{49.3322}
& 3.793 & 0.6111 & 66.0845
& \second{2.3705} & 65.7693 \\

~ & RealBasicVSR \cite{chan2022realbasicvsr}
& \best{0.2076} & \best{0.1278} & \best{48.6680}
& \best{3.292} & 0.6249 & 69.4399
& 2.8075 & 69.9027 \\

~ & AdcSR \cite{chen2025adversarial}
& 0.2553 & 0.1533 & 71.9473
& 3.500 & \third{0.6382} & \best{72.0246}
& 3.9597 & \third{71.5490} \\

~ & UpscaleAVideo \cite{zhou2024upscale}
& 0.3420 & 0.1995 & 95.4628
& 4.821 & 0.4886 & 53.8774
& \third{2.7397} & 52.0868 \\

~ & DOVE \cite{chendove}
& \third{0.2192} & \second{0.1287} & \third{56.5175}
& 4.463 & 0.6320 & 70.1333
& \best{2.2717} & 48.3011 \\

~ & STAR \cite{xie2025star}
& 0.2857 & 0.1517 & 63.2338
& 4.288 & 0.6376 & 67.0929
& 3.0611 & 62.6324 \\

~ & DLoRAL \cite{sun2025one}
& 0.2494 & 0.1596 & 58.9229
& 3.615 & \second{0.6471} & \third{70.7384}
& 3.8593 & 70.5281 \\

\rowcolor{oursbg} \multirow{-9}{*}{SPMCS}
& InstaVSR (Ours)
& 0.2623 & \third{0.1475} & 59.2432
& \second{3.441} & \best{0.6779} & \second{71.3068}
& 3.3329 & \best{72.5530} \\

\midrule

~ & RealESRGAN \cite{wang2021real}
& 0.3188 & 0.1560 & 40.8991
& 3.293 & 0.5101 & 62.4595
& 8.3914 & \second{87.4597} \\

~ & RealViformer \cite{zhang2024realviformer}
& \third{0.2702} & 0.1540 & \third{23.7465}
& \third{3.273} & 0.5498 & 64.1624
& 5.1029 & 86.6196 \\

~ & RealBasicVSR \cite{chan2022realbasicvsr}
& 0.3167 & 0.1674 & 32.3203
& \best{3.051} & \third{0.5675} & \third{67.2988}
& 6.0415 & \best{88.4188} \\

~ & AdcSR \cite{chen2025adversarial}
& 0.3236 & 0.1606 & 29.8406
& \second{3.245} & 0.5671 & \second{70.4673}
& 10.2355 & 86.5633 \\

~ & UpscaleAVideo \cite{zhou2024upscale}
& 0.4331 & 0.2244 & 51.6875
& 5.900 & 0.3357 & 34.3326
& \best{3.5438} & 62.5278 \\

~ & DOVE \cite{chendove}
& \second{0.2336} & \best{0.1230} & \best{18.6305}
& 4.519 & 0.4836 & 63.0005
& \second{3.6251} & \third{87.2314} \\

~ & STAR \cite{xie2025star}
& \best{0.2121} & \third{0.1457} & 102.8143
& 6.678 & 0.3685 & 43.3237
& \third{5.1003} & 60.7029 \\

~ & DLoRAL \cite{sun2025one}
& 0.3338 & 0.1700 & 34.5033
& 3.574 & \best{0.6003} & \best{71.0752}
& 8.1360 & 86.5118 \\

\rowcolor{oursbg} \multirow{-9}{*}{YouHQ}
& InstaVSR (Ours)
& 0.3051 & \second{0.1254} & \second{20.9868}
& 3.463 & \second{0.5910} & 66.0882
& 6.7386 & 87.2016 \\

\midrule

~ & RealESRGAN \cite{wang2021real}
& 0.3576 & 0.1934 & 65.0686
& 4.074 & 0.5786 & 64.7056
& 3.7062 & 81.58 \\

~ & RealViformer \cite{zhang2024realviformer}
& \best{0.1290} & \best{0.0878} & \best{23.3183}
& 4.181 & 0.5696 & 61.8556
& \second{3.1018} & 80.06 \\

~ & RealBasicVSR \cite{chan2022realbasicvsr}
& 0.1810 & 0.1073 & 36.3774
& \second{3.988} & 0.6065 & 66.0118
& 3.1118 & 81.55 \\

~ & AdcSR \cite{chen2025adversarial}
& 0.2356 & 0.1410 & 46.5959
& \third{4.069} & \third{0.6149} & \best{70.5955}
& 4.8715 & \best{82.27} \\

~ & UpscaleAVideo \cite{zhou2024upscale}
& 0.3134 & 0.1785 & 93.0232
& 6.199 & 0.3600 & 39.4257
& 3.4148 & 57.29 \\

~ & STAR \cite{xie2025star}
& \third{0.1693} & \second{0.0898} & 38.0273
& 4.689 & 0.5425 & 61.8154
& 3.1244 & 76.15 \\

~ & DOVE \cite{chendove}
& \second{0.1518} & \third{0.0911} & \second{35.4606}
& 5.023 & 0.5078 & 62.4765
& \best{2.9801} & 78.75 \\

~ & DLoRAL \cite{sun2025one}
& 0.2279 & 0.1412 & 44.1321
& \best{3.718} & \best{0.6254} & \second{69.9747}
& 4.6991 & \third{81.67} \\

\rowcolor{oursbg} \multirow{-9}{*}{UDM10}
& InstaVSR (Ours)
& 0.2427 & 0.1051 & \third{36.2891}
& 4.466 & \second{0.6209} & \third{68.2072}
& \third{3.1096} & \second{81.75} \\

\bottomrule
\end{tabular}
}
\end{table}

\begin{table}[t]
\centering
\caption{\textbf{Efficiency comparison for generating a 30-frame
$2K\times2K$ video.}
A800 results follow the common protocol; RTX 4090 results are additionally
reported for methods that fit within 24\,GB of GPU memory during inference.
Peak memory is measured on the A800.}
\label{tab:efficiency}
\scriptsize
\setlength{\tabcolsep}{5.0pt}
\begin{tabular}{l|cccccc}
\toprule
& AdcSR & UAV & DOVE & STAR & DLoRAL & Ours \\
\midrule
Params (B)
& 0.46 & 14.44 & 10.55 & 2.49 & 1.30 & \textbf{0.45} \\
A800 time (min)
& 1.3 & 25 & 1.4 & 24.5 & 3.5 & \textbf{1.0} \\
RTX 4090 time (min)
& 1.0 & -- & -- & -- & -- & \textbf{0.77} \\
Peak memory (GB)
& 17.23 & 38.85 & 31.79 & 75.25 & 24.93 & \textbf{7.10} \\
\bottomrule
\end{tabular}
\end{table}

\subsection{Efficient One-Step Diffusion Architecture}

Standard latent diffusion models use a VAE encoder to map an image into a
compact latent space before denoising. In one-step VSR, however, the LR input
already provides strong structural information. We therefore remove the VAE
encoder from the generator input path to avoid additional learned compression
and computational overhead.

Instead, the aligned multi-frame input is transformed using Pixel
Unshuffle~\citep{shi2016real}:
\begin{equation}
    X_t
    =
    \operatorname{PixelUnshuffle}
    \left(
        \operatorname{Concat}
        [\tilde{I}_{t-k}^l,\ldots,I_t^l,\ldots,\tilde{I}_{t+k}^l],
        s
    \right),
    \label{eq:pixel_unshuffle}
\end{equation}
where $s$ is the Pixel Unshuffle factor. Pixel Unshuffle rearranges each
$s\times s$ spatial region into the channel dimension. It is an invertible
rearrangement and does not discard input values. The input stem of the UNet is
modified to accept the resulting channel dimension.

We further remove text--image cross-attention layers because InstaVSR uses no
text condition, and remove timestep embeddings because the model performs a
fixed one-step reconstruction rather than iterative diffusion sampling.
Finally, the bottleneck is reduced to a single residual block. These
simplifications reduce the parameter count and inference cost while retaining
the generative prior of the pretrained diffusion backbone. Architecture and
initialization details are given in
Appendix~\ref{app:architecture_details}.

\subsection{Dual-Space Adversarial Learning}
\label{sec:dual_adversarial}

Reconstruction and temporal losses encourage fidelity and consistency but
may produce conservative or oversmoothed results, especially after aggressive
backbone simplification. A single discriminator is insufficient in this
setting: latent-space supervision captures high-level image statistics but
may overlook local artifacts, whereas pixel-space supervision is sensitive to
fine textures but provides weaker semantic guidance. We therefore introduce
two complementary discriminators operating in latent and pixel spaces. Both
are used only during training and introduce no additional inference cost or test-time model parameters.

\begin{figure}[t]
  \centering
  \safeincludegraphics[width=0.96\linewidth]{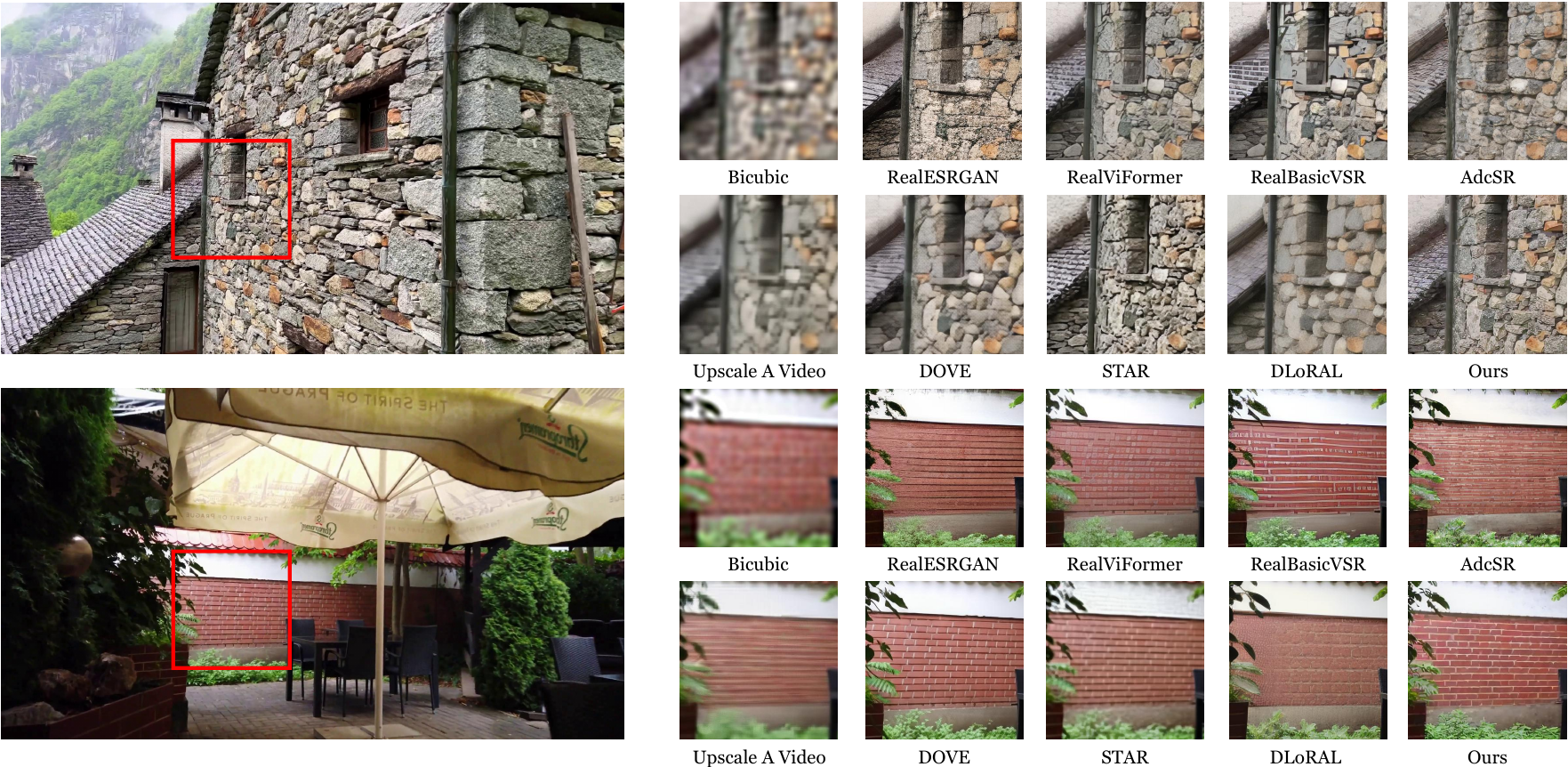}
  \caption{\textbf{Qualitative comparison of InstaVSR and other VSR methods.}
  InstaVSR reconstructs realistic textures and regular structures while
  avoiding the oversmoothing and distorted patterns observed in competing
  methods. Zoom in for the best view.}
  \label{fig:4}
\end{figure}

\paragraph{Latent-space discrimination.}

A frozen SD2.1 VAE encoder $\mathcal{E}$ maps generated and real HR frames
into the same latent space:
\begin{equation}
    z_{\mathrm{gen}}
    =
    \mathcal{E}(\hat{I}),
    \qquad
    z_{\mathrm{real}}
    =
    \mathcal{E}(I^{\mathrm{GT}}).
    \label{eq:latent_encoding}
\end{equation}
Only the VAE encoder is used in this branch. The VAE decoder is unnecessary
because both inputs are already RGB images and the purpose of
$\mathcal{E}$ is only to obtain their latent representations.

We reuse the pretrained SD2.1 UNet as the latent discriminator
$D_{\mathrm{latent}}$. Its original parameters are frozen, and only
lightweight LoRA modules~\citep{hu2022lora} are optimized. LoRA allows the
discriminator to adapt to artifacts produced by the pruned VSR generator
without the computational cost and potential prior degradation of fully
fine-tuning the UNet.

\paragraph{Pixel-space discrimination.}
Latent-space supervision may overlook small artifacts weakened by VAE
compression. We therefore introduce a pixel discriminator
$D_{\mathrm{pixel}}$ based on a frozen DINOv3-Large
backbone~\citep{simeoni2025dinov3} and lightweight trainable discrimination
heads. Operating on RGB frames allows it to detect local defects such as
jagged edges, blocky structures, unnatural textures, and residual high-frequency image noise.

The generator-side dual-space adversarial objective is
\begin{equation}
    \mathcal{L}_{G}^{\mathrm{dual}}
    =
    \lambda_{\mathrm{latent}}
    \mathcal{L}_{\mathrm{adv}}^{\mathrm{latent}}
    +
    \lambda_{\mathrm{pixel}}
    \mathcal{L}_{\mathrm{adv}}^{\mathrm{pixel}}.
    \label{eq:dual_adversarial_loss}
\end{equation}
The complete generator objective is
\begin{equation}
    \mathcal{L}_{G}
    =
    \mathcal{L}_{\mathrm{rec}}
    +
    \lambda_{\mathrm{temp}}
    \mathcal{L}_{\mathrm{temp}}
    +
    \mathcal{L}_{G}^{\mathrm{dual}}.
    \label{eq:overall_generator_loss}
\end{equation}
Latent discriminator preserves high-level generative statistics, while
the pixel discriminator improves local texture realism. Their complete
training objectives are provided in
Appendix~\ref{app:adversarial_details}.

\section{Experiments}

\subsection{Experimental Settings}
\label{sec:experimental_settings}

We train a $4\times$ VSR model for 200 epochs on a single NVIDIA H100 GPU,
using an initial learning rate of $2\times10^{-5}$, halved every 50 epochs,
and a batch size of 2. The generator is initialized from the
same-name and same-shape parameters of AdcSR~\cite{chen2025adversarial}.
Each sample contains seven consecutive frames, from which five-frame windows
are constructed for recurrent prediction. LQ--GT pairs are generated with
the RealESRGAN degradation pipeline, using shared degradation parameters
within each sequence. A lightweight PWC-Net-distilled flow network is used
for alignment and temporal supervision across neighboring video frames.

The training set contains approximately 20,000 YouTube videos at resolutions
from 720p to 2K. We evaluate on SPMCS~\cite{tao2017detail},
YouHQ~\cite{zhou2024upscale}, and UDM10 using LPIPS, DISTS~\cite{ding2020image},
FID~\cite{heusel2017gans}, NIQE~\cite{mittal2012making}, MANIQA~\cite{yang2022maniqa},
MUSIQ~\cite{ke2021musiq}, $E_{\mathrm{warp}}$, and DOVER~\cite{wu2023exploring}.
Unless otherwise specified, efficiency is measured by generating a 30-frame
$2K\times2K$ video on an NVIDIA A800. RTX 4090 results are reported only for
methods fitting within its 24\,GB memory budget.

\subsection{Quantitative Results}

\begin{figure}[t]
  \centering
  \safeincludegraphics[width=\linewidth]{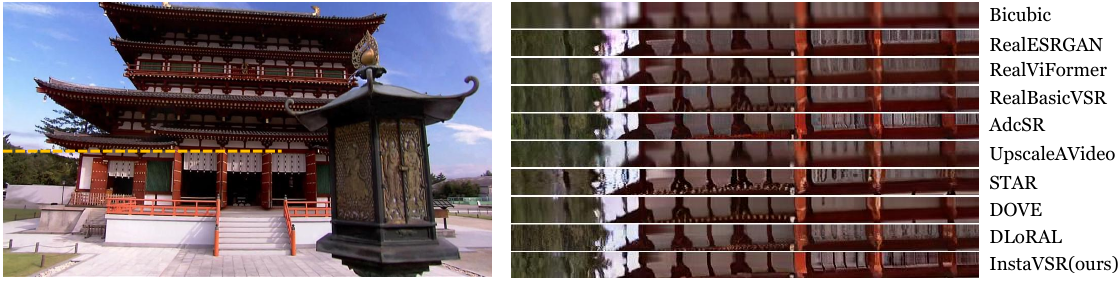}
  \caption{\textbf{Temporal profiles of video reconstruction.}
  InstaVSR produces smoother profiles while preserving straight lines and
  fine structural details across frames. Zoom in for the best view.}
  \label{fig:5}
\end{figure}

\begin{figure}[t]
    \centering
    \resizebox{\linewidth}{!}{%
        \begin{tabular}{@{}c@{\hspace{3mm}}c@{}}
            \includegraphics[height=0.14\textheight]{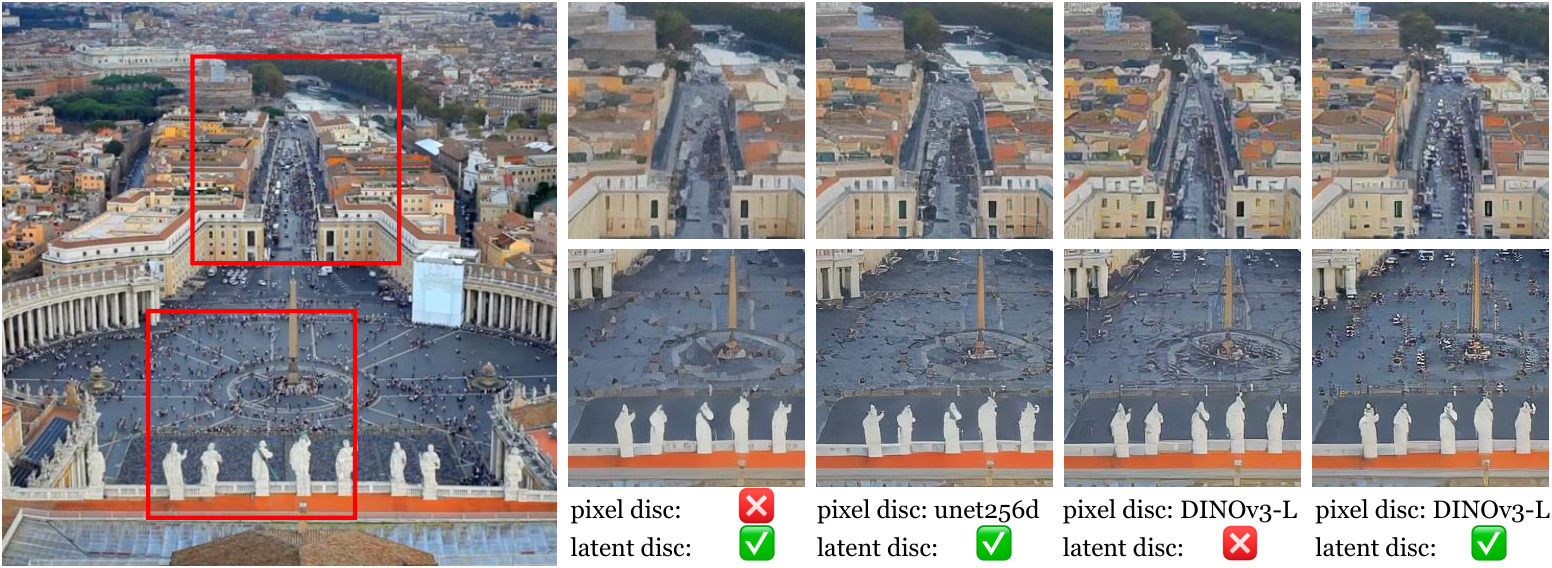}
            &
            \includegraphics[height=0.14\textheight]{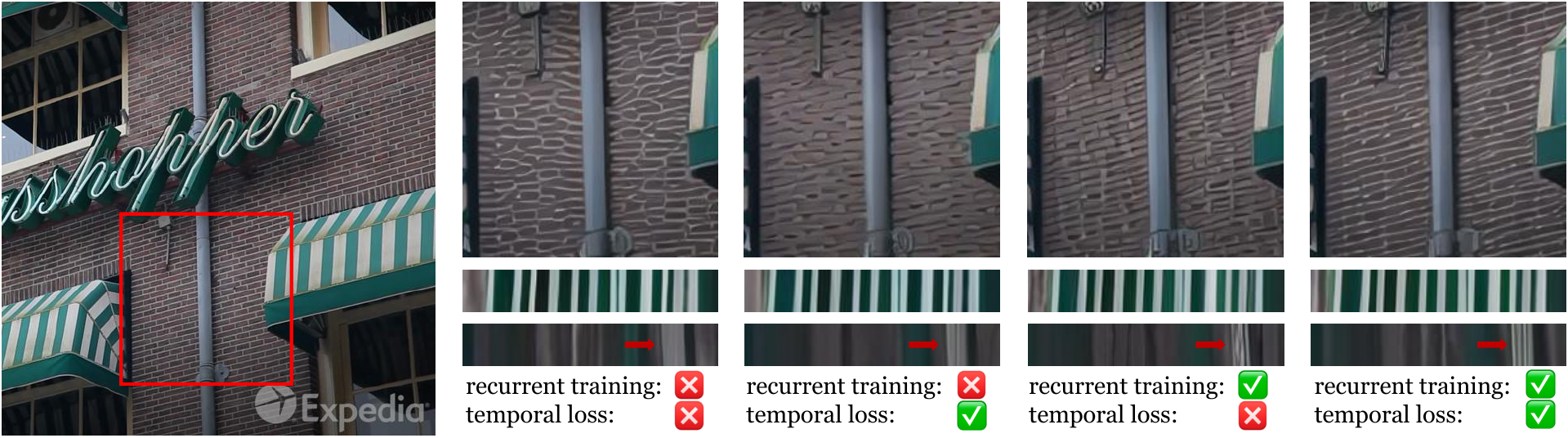}
            \\[1mm]
            {\scriptsize (a) Dual-space adversarial learning}
            &
            {\scriptsize (b) Recurrent learning and temporal regularization}
        \end{tabular}%
    }
    \caption{\textbf{Qualitative results of the ablation study.} Zoom in for the best view.}
    \label{fig:ablation_qualitative}
\end{figure}

Table~\ref{tab:main_results} shows that InstaVSR achieves a favorable
perception--fidelity--temporal trade-off across all three benchmarks.
On SPMCS, it obtains the best MANIQA and DOVER scores, the second-best
NIQE and MUSIQ scores, and the third-best DISTS score.
On YouHQ, InstaVSR ranks second in DISTS, FID, and MANIQA, while remaining
competitive in DOVER.
On UDM10, it ranks second in MANIQA and DOVER and third in FID, MUSIQ,
and $E_{\mathrm{warp}}$.
These results indicate that InstaVSR produces perceptually
strong videos with competitive overall video quality, although regression-oriented methods
retain an advantage on several distortion metrics.

As shown in Table~\ref{tab:efficiency}, InstaVSR contains only 0.45B
parameters and requires 7.10\,GB of peak GPU memory, both the lowest among
the compared methods.
It processes a 30-frame $(2K)^{2}$ video in 1.0 minute on an A800 and
0.77 minute on an RTX 4090.
The lightweight flow estimator costs 4.821\,ms per frame, corresponding
to approximately 0.31\% of the complete RTX 4090 runtime.
Thus, flow estimation accounts for only a small fraction of the measured
runtime while supporting aligned multi-frame reconstruction across neighboring low-resolution video frames.

\begin{table}[t]
\centering
\caption{\textbf{Component ablation study on SPMCS.}
Results report differences from the matched Full configuration,
$\Delta=\text{Variant}-\text{Full}$.
\textcolor{teal}{Teal} and \textcolor{red!70!black}{red}
indicate improvement and degradation, respectively.}
\label{tab:expanded_ablation}

\begingroup
\fontsize{5.5}{5.9}\selectfont
\setlength{\tabcolsep}{1.45pt}
\renewcommand{\arraystretch}{1}
\setlength{\aboverulesep}{0.15ex}
\setlength{\belowrulesep}{0.15ex}

\resizebox{0.9\textwidth}{!}{
\begin{tabular}{l|rrrrrrr}
\toprule
Configuration
& PSNR$\uparrow$
& SSIM$\uparrow$
& NIQE$\downarrow$
& MANIQA$\uparrow$
& CLIPIQA$\uparrow$
& MUSIQ$\uparrow$
& DOVER$\uparrow$ \\
\midrule

Full (reference)
& 21.764 & 0.557 & 3.441 & 0.678 & 0.609 & 71.307 & 72.553 \\

\midrule
\multicolumn{8}{l}{\textit{Dual-space adversarial learning}} \\

w/o $\mathcal{D}_{\mathrm{latent}}$
& \ablgain{+0.268}
& \ablgain{+0.016}
& \ablloss{+0.196}
& \ablgain{+0.003}
& \ablgain{+0.006}
& \ablgain{+0.108}
& \ablloss{$-$0.255} \\

w/o $\mathcal{D}_{\mathrm{pixel}}$
& \ablgain{+0.887}
& \ablgain{+0.039}
& \ablloss{+0.692}
& \ablloss{$-$0.039}
& \ablloss{$-$0.014}
& \ablloss{$-$2.674}
& \ablloss{$-$6.164} \\

\midrule
\multicolumn{8}{l}{\textit{Recurrent learning and temporal regularization}} \\

First-order recurrence
& \ablgain{+0.254}
& \ablgain{+0.013}
& \ablloss{+0.135}
& \ablloss{$-$0.006}
& \ablloss{$-$0.004}
& \ablloss{$-$0.929}
& \ablloss{$-$1.945} \\

w/o recurrent learning
& \ablgain{+0.150}
& \ablgain{+0.013}
& \ablloss{+0.288}
& \ablgain{+0.015}
& \ablgain{+0.038}
& \ablgain{+0.130}
& \ablloss{$-$0.155} \\

w/o temporal loss
& \ablgain{+0.255}
& \ablgain{+0.014}
& \ablloss{+0.046}
& \ablloss{$-$0.008}
& \ablloss{$-$0.007}
& \ablloss{$-$0.929}
& \ablloss{$-$2.014} \\

w/o recurrent and temporal loss
& \ablgain{+0.257}
& \ablgain{+0.014}
& \ablloss{+0.342}
& \ablloss{$-$0.021}
& \ablloss{$-$0.058}
& \ablloss{$-$4.266}
& \ablloss{$-$7.544} \\


\midrule
\multicolumn{8}{l}{\textit{Flow-guided temporal alignment}} \\

w/o flow alignment
& \ablgain{+0.578}
& \ablgain{+0.025}
& \ablloss{+0.350}
& \ablloss{$-$0.058}
& \ablloss{$-$0.087}
& \ablloss{$-$5.771}
& \ablloss{$-$10.732} \\


\midrule
\multicolumn{8}{l}{\textit{One-step backbone design}} \\

w/ VAE encoder
& \ablgain{+0.260}
& \ablgain{+0.026}
& \ablgain{$-$0.071}
& \ablgain{+0.008}
& \ablloss{$-$0.021}
& \ablgain{+0.420}
& \ablgain{+0.160} \\

w/ cross-attention and time embedding
& \ablgain{+0.090}
& \ablgain{+0.013}
& \ablgain{$-$0.034}
& \ablgain{+0.005}
& \ablloss{$-$0.018}
& \ablgain{+0.310}
& \ablloss{$-$0.120} \\






\bottomrule
\end{tabular}
}
\endgroup
\end{table}

\subsection{Qualitative Results}

Figure~\ref{fig:4} compares spatial restoration quality on real-world
videos.
InstaVSR recovers fine-grained stone, brick, foliage, and soil textures
while preserving the underlying scene structure.
In contrast, competing methods either oversmooth these regions or generate
irregular and geometrically inconsistent patterns in the illustrated image regions.

Figure~\ref{fig:5} presents temporal profiles formed by stacking the same
horizontal scanline from consecutive frames.
InstaVSR produces smooth trajectories with well-preserved structural lines,
whereas competing methods either remove fine structures or introduce jagged
and temporally varying patterns.
This demonstrates that the proposed method balances temporal smoothness
with geometric fidelity in the illustrated temporal profiles.

\subsection{Ablation Study}

\paragraph{Adversarial supervision.}
Removing the pixel discriminator substantially reduces MANIQA, CLIPIQA,
MUSIQ, and DOVER, despite improving distortion-oriented metrics.
Removing the latent discriminator produces smaller and mixed changes,
indicating that the two discriminators provide complementary local and
semantic supervision.

\paragraph{Recurrent and flow-guided learning.}
First-order recurrence underperforms the complete second-order formulation
on perceptual and video-quality metrics.
Removing both recurrent learning and temporal loss causes the largest
degradation in this group, reducing MUSIQ and DOVER by 4.266 and 7.544,
respectively.
The particularly large DOVER decrease without flow alignment supports the
use of motion compensation when exploiting neighboring observations.
The occlusion mask further prevents unreliable correspondences from
dominating temporal supervision.

\paragraph{Backbone design.}
Restoring the VAE encoder or cross-attention and timestep embeddings produces
only mixed and modest quality changes, motivating their removal in our
efficiency-oriented design. Temporal context size is analyzed in Appendix~\ref{app:context_size}.
The qualitative results in Figure~\ref{fig:ablation_qualitative} support
the quantitative observations.
Without the pixel discriminator, outputs exhibit oversmoothed or blocky
textures, while removing latent-space supervision increases
content-inconsistent details.
Without recurrent learning and temporal regularization, structures drift
under repeated prediction and temporal variations become more visible.
The complete configuration provides a favorable balance between realistic
detail, structural fidelity, and temporal stability.

\section{Conclusion}

We presented InstaVSR, a lightweight one-step diffusion framework that
balances spatial efficiency and temporal consistency for video
super-resolution. Through systematic component analysis, we integrate
a pruned backbone, flow-aligned multi-frame inputs, and dual-space
adversarial supervision to reduce computational cost while preserving
perceptual quality. Our recurrent second-order learning writes SR
predictions back into the observation buffer and revisits target frames
under an SR-enhanced context. Together with flow-guided temporal
regularization, this promotes stable reconstruction without heavy
spatiotemporal architectures. Experiments on three benchmarks demonstrate
a favorable balance among perceptual quality, temporal consistency, and
efficiency. With 0.45B parameters, InstaVSR processes a 30-frame
$2K\times2K$ sequence in 0.77 minutes on an RTX 4090 under the reported inference protocol.

Despite these gains, InstaVSR does not consistently outperform
regression-oriented methods on fidelity metrics. Future work will
investigate how to improve reconstruction fidelity while retaining
perceptual quality and temporal stability under a limited computational
budget.

\clearpage
\section*{Reproducibility Statement}
Section~\ref{sec:experimental_settings} reports the training schedule, model
initialization, datasets, and evaluation hardware.
Appendix~\ref{app:recurrent_details} details recurrent prediction and temporal
supervision; Appendices~\ref{app:architecture_details}
and~\ref{app:adversarial_details} specify the architecture, initialization,
and adversarial objectives. Data preparation and evaluation metrics are
described in Appendices~\ref{app:data_protocol} and~\ref{app:metrics}.
The experimental tables state the reference configurations and distinguish
benchmark-resolution latency from the $2K\times2K$ efficiency protocol.

\section*{AI Use Statement}
Generative AI assistance was limited to language polishing and manuscript
formatting. The authors remain responsible for verifying the revisions and
for the scientific claims, experimental results, and final content of this work.

\bibliographystyle{iclr2027_conference}
\bibliography{references}

\clearpage
\appendix

\section{Additional Methodological Details}
\label{app:method_details}

\subsection{Recurrent Prediction and Temporal Regularization}
\label{app:recurrent_details}

\paragraph{Seven-frame training sequence.}
We sample seven consecutive LR frames
$\{I_{t-3}^l,\ldots,I_{t+3}^l\}$ and initialize
$B_j^{[0]}=I_j^l$. The buffer is updated by sequentially
reconstructing frames $t-1$, $t$, and $t+1$:
\begin{equation}
\begin{aligned}
    \hat{I}_{t-1}^{h,(1)}
    &=
    G_\theta\!\left(
        \mathcal{A}_{t-1}(B_{t-3:t+1}^{[0]})
    \right), \\
    B_{t-1}^{[1]}
    &=
    \mathcal{R}\!\left(\hat{I}_{t-1}^{h,(1)}\right),
\end{aligned}
\end{equation}
\begin{equation}
\begin{aligned}
    \hat{I}_{t}^{h,(1)}
    &=
    G_\theta\!\left(
        \mathcal{A}_{t}(B_{t-2:t+2}^{[1]})
    \right), \\
    B_{t}^{[2]}
    &=
    \mathcal{R}\!\left(\hat{I}_{t}^{h,(1)}\right),
\end{aligned}
\end{equation}
and
\begin{equation}
\begin{aligned}
    \hat{I}_{t+1}^{h,(1)}
    &=
    G_\theta\!\left(
        \mathcal{A}_{t+1}(B_{t-1:t+3}^{[2]})
    \right), \\
    B_{t+1}^{[3]}
    &=
    \mathcal{R}\!\left(\hat{I}_{t+1}^{h,(1)}\right).
\end{aligned}
\end{equation}
At each update, all buffer entries other than the reconstructed
center frame remain unchanged. Thus, the first prediction of
frame $t$ already uses the SR result of frame $t-1$, and the
first prediction of frame $t+1$ uses both preceding SR results.

We then reconstruct frame $t$ again using the updated buffer:
\begin{equation}
    \hat{I}_{t}^{h,(2)}
    =
    G_\theta\!\left(
        \mathcal{A}_{t}(B_{t-2:t+2}^{[3]})
    \right).
\end{equation}
This second-order prediction uses a window containing the
first-order SR results of frames $t-1$, $t$, and $t+1$.
These frames correspond to frames 2, 3, and 4 in
Figure~\ref{fig:3}.

\paragraph{Supervision.}
The reconstruction loss in Eq.~\ref{eq:reconstruction_loss}
compares $\hat{I}_{t+1}^{h,(1)}$ with its HR target $I_{t+1}^h$.
The temporal loss in Eq.~\ref{eq:temporal_loss} compares
$\hat{I}_{t}^{h,(2)}$ with $\hat{I}_{t+1}^{h,(1)}$ after
flow alignment. The recurrent process constructs the
first- and second-order predictions used for this supervision;
it does not introduce a separate rollout loss during the recurrent training process.

\paragraph{Flow processing.}
The optical-flow estimator operates on degraded LR inputs.
For an upscaling factor $r$, the estimated flow is resized
to the HR spatial resolution, and both displacement components
are multiplied by $r$ before warping the SR predictions.

Let $F_{t\rightarrow t+1}$ denote the resulting flow used
to warp the second-order prediction of frame $t$ into the
coordinate system of frame $t+1$. Forward--backward flow
consistency provides a visibility mask $M_{t\rightarrow t+1}$,
which excludes occluded regions and unreliable correspondences.
The resulting per-pixel confidence weight is defined as follows:
\begin{equation}
    \omega_{t\rightarrow t+1}(p)
    =
    M_{t\rightarrow t+1}(p)
    \exp\!\left(
        -\frac{
            \|F_{t\rightarrow t+1}(p)\|_2
        }{\sigma_m}
    \right),
\end{equation}
where $\sigma_m$ controls the weighting of motion magnitude.
This reduces the contribution of large-motion regions, where
flow estimation may be less reliable.

The weighted temporal loss penalizes differences between
the aligned second-order and first-order predictions,
while reconstruction supervision anchors the first-order
prediction to its HR target.

\subsection{One-Step Architecture and Initialization}
\label{app:architecture_details}

\paragraph{Pixel Unshuffle input transformation.}
For $K=2k+1$ aligned RGB frames, concatenation produces an input with
$3K$ channels:
\begin{equation}
    X_t^{\mathrm{cat}}
    =
    \operatorname{Concat}
    [\tilde{I}_{t-k}^l,\ldots,I_t^l,\ldots,\tilde{I}_{t+k}^l].
\end{equation}
Given a Pixel Unshuffle factor $s$, the transformation maps
\begin{equation}
    X_t^{\mathrm{cat}}
    \in
    \mathbb{R}^{H_l\times W_l\times 3K}
\end{equation}
to
\begin{equation}
    X_t
    \in
    \mathbb{R}^{
        \frac{H_l}{s}
        \times
        \frac{W_l}{s}
        \times
        3Ks^2
    }.
\end{equation}
Pixel Unshuffle changes only the arrangement of the input values and is
therefore invertible. Unlike a learned VAE encoder, it does not introduce
additional nonlinear compression before processing with the one-step diffusion UNet.

The input convolution of the UNet is modified to accept the expanded channel
dimension. The aligned frames share the same target coordinate system before
Pixel Unshuffle, allowing the input stem to jointly process their spatial and
temporal information.

\paragraph{Backbone pruning.}
The original diffusion backbone contains text--image cross-attention layers
for injecting text embeddings and timestep-embedding modules for representing
different denoising steps. Neither component is required in InstaVSR because
the aligned LR frames provide the conditioning signal and the generator
performs a fixed one-step reconstruction.

We therefore remove the text--image cross-attention layers and timestep
embeddings. We also reduce the UNet bottleneck to a single residual block.
These modifications decrease the parameter count and computational cost while
retaining the spatial representation inherited from the pretrained diffusion
backbone used for model initialization.

The VAE is removed only from the generator input path. During training, the
latent discriminator still uses a frozen VAE encoder to map generated and
real RGB images into the pretrained latent space. No VAE decoder is used by
the latent discriminator, and neither the latent nor the pixel discriminator
is retained during inference.

\paragraph{Parameter initialization.}
The generator is initialized from AdcSR by loading parameters whose names and
tensor shapes match the corresponding InstaVSR modules. Parameters associated
with the expanded multi-frame input stem and structurally modified modules
are initialized separately and trained for InstaVSR. The recurrent buffer,
flow-guided temporal objectives, and adversarial components are also newly
introduced and are not inherited from AdcSR.

Although concatenating multiple aligned frames increases the channel
dimension of the input stem, the middle and bottleneck layers are simplified.
As a result, InstaVSR remains slightly smaller than the single-frame AdcSR
generator while exploiting multi-frame information.

\subsection{Complete Dual-Space Adversarial Objectives}
\label{app:adversarial_details}

\paragraph{Latent-space discriminator.}
Given a generated frame $\hat{I}$ and its ground-truth HR frame
$I^{\mathrm{GT}}$, the frozen SD2.1 VAE encoder $\mathcal{E}$ produces
\begin{equation}
    z_{\mathrm{gen}}
    =
    \mathcal{E}(\hat{I}),
    \qquad
    z_{\mathrm{real}}
    =
    \mathcal{E}(I^{\mathrm{GT}}).
\end{equation}
The pretrained SD2.1 UNet is reused as the latent discriminator
$D_{\mathrm{latent}}$. Its dense outputs are interpreted as patch-wise
discrimination scores and aggregated when computing the adversarial loss.
The original UNet parameters remain frozen, while trainable LoRA
modules~\citep{hu2022lora} adapt the discriminator to artifacts produced by
the VSR generator.

The latent discriminator is optimized with the non-saturating logistic loss:
\begin{equation}
\begin{aligned}
    \mathcal{L}_{D}^{\mathrm{latent}}
    ={}&
    \mathbb{E}_{I^{\mathrm{GT}}}
    \left[
        \operatorname{softplus}
        \left(
            -D_{\mathrm{latent}}(z_{\mathrm{real}})
        \right)
    \right] \\
    &+
    \mathbb{E}_{\hat{I}}
    \left[
        \operatorname{softplus}
        \left(
            D_{\mathrm{latent}}
            \left(
                \operatorname{sg}(z_{\mathrm{gen}})
            \right)
        \right)
    \right],
\end{aligned}
\label{eq:appendix_latent_discriminator}
\end{equation}
where $\operatorname{sg}(\cdot)$ stops gradients from propagating to the
generator during the discriminator update. The corresponding adversarial
objective for the generator is
\begin{equation}
    \mathcal{L}_{\mathrm{adv}}^{\mathrm{latent}}
    =
    \mathbb{E}_{\hat{I}}
    \left[
        \operatorname{softplus}
        \left(
            -D_{\mathrm{latent}}(z_{\mathrm{gen}})
        \right)
    \right].
    \label{eq:appendix_latent_generator}
\end{equation}

\paragraph{Pixel-space discriminator.}
The pixel discriminator $D_{\mathrm{pixel}}$ uses a frozen DINOv3-Large
backbone~\citep{simeoni2025dinov3}. Features extracted from three selected
backbone levels are passed to lightweight trainable ConvNeXt heads, producing
patch-wise real/fake predictions at multiple feature levels. The DINOv3
backbone remains frozen and in evaluation mode, while only the discriminator
heads are optimized. DiffAugment is applied consistently to real and generated
images during discriminator training at each optimization step.

The pixel discriminator is trained with the hinge loss:
\begin{equation}
\begin{aligned}
    \mathcal{L}_{D}^{\mathrm{pixel}}
    ={}&
    \mathbb{E}_{I^{\mathrm{GT}}}
    \left[
        \max
        \left(
            0,\,
            1-D_{\mathrm{pixel}}(I^{\mathrm{GT}})
        \right)
    \right] \\
    &+
    \mathbb{E}_{\hat{I}}
    \left[
        \max
        \left(
            0,\,
            1+
            D_{\mathrm{pixel}}
            \left(
                \operatorname{sg}(\hat{I})
            \right)
        \right)
    \right].
\end{aligned}
\label{eq:appendix_pixel_discriminator}
\end{equation}
The corresponding generator objective is
\begin{equation}
    \mathcal{L}_{\mathrm{adv}}^{\mathrm{pixel}}
    =
    -
    \mathbb{E}_{\hat{I}}
    \left[
        D_{\mathrm{pixel}}(\hat{I})
    \right].
    \label{eq:appendix_pixel_generator}
\end{equation}

The complete dual-space adversarial objective is
\begin{equation}
    \mathcal{L}_{G}^{\mathrm{dual}}
    =
    \lambda_{\mathrm{latent}}
    \mathcal{L}_{\mathrm{adv}}^{\mathrm{latent}}
    +
    \lambda_{\mathrm{pixel}}
    \mathcal{L}_{\mathrm{adv}}^{\mathrm{pixel}}.
\end{equation}
Combining reconstruction, flow-guided temporal supervision,
and dual-space adversarial learning yields the following complete generator training objective:
\begin{equation}
    \mathcal{L}_{G}
    =
    \mathcal{L}_{\mathrm{rec}}
    +
    \lambda_{\mathrm{temp}}
    \mathcal{L}_{\mathrm{temp}}
    +
    \lambda_{\mathrm{latent}}
    \mathcal{L}_{\mathrm{adv}}^{\mathrm{latent}}
    +
    \lambda_{\mathrm{pixel}}
    \mathcal{L}_{\mathrm{adv}}^{\mathrm{pixel}}.
\end{equation}

The latent discriminator encourages the generated frames to follow the
high-level statistics represented by the pretrained diffusion model, whereas
the pixel discriminator penalizes local RGB artifacts that may be weakened
by VAE encoding. Since both discriminators are removed after training, the
dual-space objective does not increase inference parameters or runtime.

\subsection{Training Data and Evaluation Protocol}
\label{app:data_protocol}

We collect approximately 20,000 training videos from YouTube, with
resolutions ranging from 720p to 2K.
Before generating training pairs, the source videos are enhanced using a
$1\times$ restoration model to reduce compression artifacts in the
pseudo-ground-truth data.
LQ sequences are then synthesized using the RealESRGAN degradation pipeline.
The degradation parameters are shared across all frames of each sequence,
which avoids unrealistic frame-wise changes in blur, noise, and compression across neighboring video frames.

In addition to the three quantitative benchmarks, we collect a real-world
aerial-video set for qualitative analysis.
It contains 10 scenes at 2K resolution, with each sequence consisting of
30 frames captured at 30 fps.
The frame rate describes the source-video acquisition setting and does not
imply real-time inference by the proposed model.

Training and inference hardware are reported separately.
The model is trained on one NVIDIA H100 GPU, whereas the main efficiency
comparison uses an NVIDIA A800 under the same 30-frame
$2K\times2K$ protocol.
RTX 4090 measurements are additionally provided for methods that fit within
24\,GB of memory.
The use of an H100 for training therefore does not affect the fairness of
the reported comparison of inference efficiency across methods.

\subsection{Evaluation Metrics}
\label{app:metrics}

We evaluate spatial fidelity using LPIPS, DISTS, and FID, and assess
no-reference perceptual quality using NIQE, MANIQA, CLIPIQA, and MUSIQ.
We report DOVER for overall video quality and use the following flow-guided
warping error to quantify motion-compensated temporal consistency:

\begin{equation}
E_{\mathrm{warp}}
=
\frac{1}{T-1}
\sum_{t=1}^{T-1}
\left\|
\mathcal{W}
\left(
\hat{I}_{t+1},
F_{t+1\rightarrow t}
\right)
-
\hat{I}_{t}
\right\|_{1},
\end{equation}

where $\mathcal{W}$ denotes backward warping and
$F_{t+1\rightarrow t}$ is the estimated optical flow from frame $t+1$
to frame $t$.
A lower $E_{\mathrm{warp}}$ indicates stronger motion-compensated temporal
consistency.
DOVER complements this low-level measure by assessing overall perceptual
video quality.

\subsection{Comparison with RVRT}

\begin{table}[t]
\centering
\caption{\textbf{Comparison with the efficient non-diffusion baseline RVRT
on benchmark-resolution SPMCS clips.}
Latency is measured over 30 frames and is not directly comparable with the
$2K\times2K$ protocol used in the main paper.}
\label{tab:rvrt_comparison}
\footnotesize
\setlength{\tabcolsep}{2.4pt}
\renewcommand{\arraystretch}{1}
\resizebox{\linewidth}{!}{
\begin{tabular}{l|cc|cccc|cc|c}
\toprule
Method
& LPIPS$\downarrow$
& DISTS$\downarrow$
& NIQE$\downarrow$
& MANIQA$\uparrow$
& CLIPIQA$\uparrow$
& MUSIQ$\uparrow$
& $E_{\mathrm{warp}}\downarrow$
& DOVER$\uparrow$
& \makecell{Latency$\downarrow$\\(s/30 frames)} \\
\midrule
RVRT \cite{liang2022recurrent}
& 0.1835 & 0.1028
& 5.1069 & 0.6304 & 0.4476 & 63.2708
& 1.4734 & 61.3796 & 18.312 \\

InstaVSR
& 0.2623 & 0.1475
& 3.4410 & 0.6779 & 0.6093 & 71.3068
& 3.3329 & 72.5530 & 8.167 \\
\bottomrule
\end{tabular}
}
\end{table}

Table~\ref{tab:rvrt_comparison} compares InstaVSR with RVRT, an efficient
non-diffusion VSR method.
RVRT performs better on LPIPS, DISTS, and warping error,
whereas InstaVSR achieves substantially stronger NIQE, MANIQA, CLIPIQA,
MUSIQ, and DOVER scores.
Under this benchmark-resolution protocol, InstaVSR also requires less than
half the measured latency.
The results reflect the different objectives of deterministic regression
and generative restoration: RVRT prioritizes pixel fidelity, while InstaVSR
provides stronger perceptual quality and overall video quality.


\subsection{Temporal Context Size}
\label{app:context_size}

\begin{table}[t]
\centering
\caption{\textbf{Temporal context size ablation on SPMCS.}
The five-frame configuration reports absolute values; three- and
seven-frame configurations report differences from this reference,
$\Delta=\text{Variant}-\text{5-frame}$.
\textcolor{teal}{Teal} and \textcolor{red!70!black}{dark red}
indicate improvement and degradation, respectively.}
\label{tab:context_size}
\small
\setlength{\tabcolsep}{8pt}
\renewcommand{\arraystretch}{1}
\begin{tabular}{l|cccc}
\toprule
Setting
& \makecell{Flow time\\(ms/frame)$\downarrow$}
& \makecell{Total time\\(s/frame)$\downarrow$}
& MANIQA$\uparrow$
& $E_{\mathrm{warp}}\downarrow$ \\
\midrule
5-frame (Ours)
& 4.821
& 1.54
& 0.678
& 3.333 \\

3-frame
& \ablgain{$-$1.070}
& \ablgain{$-$0.11}
& \ablloss{$-$0.017}
& \ablloss{+0.194} \\

7-frame
& \ablloss{+1.088}
& \ablloss{+0.14}
& \ablloss{$-$0.030}
& \ablgain{$-$0.305} \\
\bottomrule
\end{tabular}
\end{table}

Table~\ref{tab:context_size} compares three-, five-, and seven-frame
input windows. Three frames reduce runtime but yield worse MANIQA
and warping error, whereas seven frames improve warping error at
the cost of runtime and perceptual quality. We therefore use five
frames as the default balance.

For the five-frame configuration, the lightweight flow estimator
costs 4.821\,ms per frame, approximately 0.31\% of the complete
RTX 4090 runtime, indicating a small flow-estimation overhead.

\subsection{Temporal Range and Occlusion Masking}
\label{app:temporal_ablation}

\begin{table}[!htbp]
\centering
\caption{\textbf{Additional temporal ablations on SPMCS.}
Results report differences from the matched Full configuration
used in Table~\ref{tab:expanded_ablation},
$\Delta=\text{Variant}-\text{Full}$.
\textcolor{teal}{Teal} and \textcolor{red!70!black}{red}
indicate improvement and degradation relative to the Full configuration, respectively.}
\label{tab:temporal_ablation}
\begingroup
\small
\setlength{\tabcolsep}{3pt}
\renewcommand{\arraystretch}{1.12}
\resizebox{\textwidth}{!}{
\begin{tabular}{l|rrrrrrrr}
\toprule
Configuration
& PSNR$\uparrow$
& SSIM$\uparrow$
& NIQE$\downarrow$
& MANIQA$\uparrow$
& CLIPIQA$\uparrow$
& MUSIQ$\uparrow$
& $E_{\mathrm{warp}}\downarrow$
& DOVER$\uparrow$ \\
\midrule

Temporal range $D=2$
& \ablgain{+0.141}
& \ablgain{+0.008}
& \ablloss{+0.438}
& \ablgain{+0.004}
& \ablloss{$-$0.014}
& \ablloss{$-$1.001}
& \ablgain{$-$0.221}
& \ablloss{$-$2.192} \\

w/o occlusion mask
& \ablloss{$-$0.018}
& 0.000
& \ablloss{+0.275}
& \ablloss{$-$0.012}
& \ablloss{$-$0.015}
& \ablloss{$-$0.948}
& \ablgain{$-$0.085}
& \ablloss{$-$4.115} \\
\bottomrule
\end{tabular}
}
\endgroup
\end{table}

Table~\ref{tab:temporal_ablation} examines temporal range and
occlusion masking. Setting $D=2$ produces mixed changes:
PSNR, SSIM, MANIQA, and warping error improve, while NIQE,
CLIPIQA, MUSIQ, and DOVER worsen. In particular, MUSIQ and
DOVER decrease by 1.001 and 2.192, respectively.

Removing the occlusion mask reduces MANIQA, CLIPIQA, MUSIQ,
and DOVER, with a DOVER drop of 4.115, despite a small
improvement in warping error. These results support retaining
occlusion masking for perceptual and overall video quality,
while showing that lower warping error alone does not
necessarily indicate better perceptual quality.

\end{document}